# Deep reinforcement learning driven inspection and maintenance planning under incomplete information and constraints

C.P. Andriotis, K.G. Papakonstantinou

Department of Civil & Environmental Engineering, The Pennsylvania State University, University Park, PA, USA

**Abstract**

Determination of inspection and maintenance policies for minimizing long-term risks and costs in deteriorating engineering environments constitutes a complex optimization problem. Major computational challenges include the (i) curse of dimensionality, due to exponential scaling of state/action set cardinalities with the number of components; (ii) curse of history, related to exponentially growing decision-trees with the number of decision-steps; (iii) presence of state uncertainties, induced by inherent environment stochasticity and variability of inspection/monitoring measurements; (iv) presence of constraints, pertaining to stochastic long-term limitations, due to resource scarcity and other infeasible/undesirable system responses. In this work, these challenges are addressed within a joint framework of constrained Partially Observable Markov Decision Processes (POMDP) and multi-agent Deep Reinforcement Learning (DRL). POMDPs optimally tackle (ii)-(iii), combining stochastic dynamic programming with Bayesian inference principles. Multi-agent DRL addresses (i), through deep function parametrizations and decentralized control assumptions. Challenge (iv) is herein handled through proper state augmentation and Lagrangian relaxation, with emphasis on life-cycle risk-based constraints and budget limitations. The underlying algorithmic steps are provided, and the proposed framework is found to outperform well-established policy baselines and facilitate adept prescription of inspection and intervention actions, in cases where decisions must be made in the most resource- and risk-aware manner.

## 1. Introduction

Optimal inspection and maintenance planning delineates a class of important engineering decision-making problems, aimed at supporting sustainable and resilient operation of systems and networks over their life-cycle. Optimality refers to minimizing various societal, environmental, and economic risks, along with other operational costs, as these emerge due to the combined consequences of the selected actions of the decision-maker and their effects on the future exogenous deterioration of the environment. Within this context, the goal of the decision-maker is to determine an appropriate policy, i.e. an optimal rule of sequential decisions over a presumed time frame, which is able to aptly map states and times to intervention and observation actions [1, 2].

Literature indicates several approaches to solving this problem, from threshold-based nonlinear and mixed-integer programming formulations (e.g. in [3, 4, 5, 6]), to analysis of decision trees (e.g. in [7, 8, 9, 10]), and from renewal theory (e.g. in [11, 12, 13, 14]), to stochastic optimal control (e.g. in [15, 16, 17, 18]). These approaches are also applicable to infrastructure problems beyond inspection and maintenance planning, such as post-disaster recovery, e.g. in [19, 20, 21]. Respectively, admissible solution strategies to the above approaches span from exhaustive policy enumeration, and genetic algorithms, to gradient-based schemes, and dynamic programming. Besides formulations that leverage dynamic programming and stochastic optimal control concepts, a common characteristic underlying traditional inspection and maintenance planning methods is that the decision-making problem, despite its inherent sequential and dynamic nature, is articulated by means of static optimization formulations. As a result, many otherwise practical approaches tend to be more susceptible to optimality limitations, especially in problems with high-dimensional spaces and long decision horizons, challenges also known as the *curse of dimensionality* and *curse of history*, respectively [22, 23]. Moreover, many solution techniques often lack cohesive and generalizable mathematical capabilities regarding the consistent integration of stochastic environments and/or uncertain observation outcomes in the optimization process, as well as the incorporation of stochastic or deterministic constraints that need to be satisfied over multiple time steps or even the entire operating life of the system.

To address the above issues, this work follows a stochastic optimal control approach, casting the optimization problem within the joint context of constrained Partially Observable Markov Decision Processes (POMDPs) and multi-agent Deep Reinforcement Learning (DRL). POMDPs are able to alleviate the curse of history as a result of their dynamic programming principles, and to facilitate optimal reasoning in the presence of real-time noisy observations [24]. Their efficiency in inspection and maintenance planning has been thoroughly studied and exemplified in [25, 26, 27, 28], among others. Within the same class of applications, in the confluence of DRL and point-based POMDPs, the Deep Centralized Multi-agent Actor Critic (DCMAC) approach has been recently developed in [29, 30], an off-policy algorithm with experience replay, belonging in the general family of actor-critic approaches [31, 32]. DCMAC leverages the concept of belief-state MDPs, a fundamental idea for the development of point-based POMDP algorithms, thus directly operating on the posterior probabilities of system states given past actions and observations [33]. In DCMAC, individual control units are centralized in terms of global state information and sharing of policy network parameters, nonetheless, they are *decentralized* in terms of policy outputs. Hence, based on classic Markov decision processes formalism, DCMAC provides Decentralized POMDP (Dec-POMDP) solutions [34, 35], for a setting where the agents representing the various control units have access to the entire state distribution of the system, however, having the autonomy to make their own choices without being aware of each other's actions. DRL is extremely efficient in tackling the curse of dimensionality stemming from high-dimensional and/or combinatoric state spaces, whereas the computational hurdle of exponential scaling of the number of actions with the number of components, is seamlessly handled by the decentralized multi-agent formulation of the problem, given that decentralization enables linear scaling.

Building upon the above described DRL concepts in this work, a modified architecture compared to the original DCMAC approach is implemented for the actor. We consider a sparser parametrization of the actor, without parameter sharing, i.e. each agent has its own individual policy network. We call this architecture Deep Decentralized Multi-agent Actor Critic (DDMAC). Similar approaches exist for various cooperative/competitive multi-agent robotic and gaming control tasks [36, 37]. Thorough reviews on state-of-the-art methods and applications can be also found in [38, 39]. Despite the architectural differences with DCMAC, DDMAC solves the same Dec-POMDP problem, eliminating, however, inter-agent interactions in the hidden layers for the sake of computational efficacy. Based on this numerical approach, this paper is particularly focused on investigating the effects of incorporating resource constraints and other limitations, especially in the forms of *budget* and *life-cycle risk constraints*. Depending on the nature of the modeled limitations, the constraints can be addressed through either state augmentation or primal-dual optimization approaches based on the Lagrangian function of the problem.

Constrained static optimization formulations for operation and maintenance policies exist in the literature, e.g. in [3, 14, 40, 41], mainly reflecting short-term risk, reliability-based, and budget-related considerations. In the case of POMDPs, the optimization problem now falls in the category of constrained POMDPs. Constrained Markov decision processes have been given model-based solutions with the aid of linear programming formulations in [42, 43]. Exact POMDP alpha-vector value interation can be extended to constrained problems as well, inheriting, however, the PSPACE complexity of the unconstrained solution [44]. Unconstrained point-based POMDPs algorithms, which are well-suited for inspection and maintenance planning of systems with up to thousands of states and hundreds of actions and observations [27, 18], have also been extended to constrained problems [45]. In multi-component systems, under the assumption of component-wise independent cost functions, states, and actions, [46] derives constrained POMDP solutions through a series of unconstrained solutions controlled by a linear master program. Overall, and notwithstanding their principled mathematical descriptions, the above value iteration and linear or nonlinear programming formulations are fundamentally hard to extend to high-dimensional systems that are of interest in this work.

In DRL, constraints typically refer to either the parameters of the approximated functions, or the cumulative returns related to auxiliary functions of interest [47, 48, 49]. The former methods restrain the iterate increment of the policy parameter updates to be within a trust region of the Kullback-Leibler divergence between the new and the old policy, thus preventing abrupt policy changes and, consequently, training instabilities. In such cases, optimization is typically based on surrogates of the objective and constraint functions [47]. The latter methods typically aim to protect the agent from unsafe or otherwise undesirable states and choices during training or policy deployment. To this end, the objective is optimized with the aid of primal-dual formulations, either through trust region concepts, or Lagrangian relaxation, or domain-based manual penalization [48, 50, 51]. Safe RL formulations similarly integrate risk and policy variance in the constraint functions of the problem, or directly intervene in exploration to guide training [52, 53]. Such "safety" constraints can for example pertain to the probability of failure over multiple steps and, as such, they reflect *soft* constraints, meaning that they only need to be satisfied in a probabilistic or expected sense. The satisfaction of *hard* constraints, such as budget constraints, are easier to account for in the optimization process through state augmentation. Such constraints tend to be relevant for other resource limitations as well, e.g. in cases of limited availability of operating crews, inspectors, etc. In this work, we consider and study both types of constraints.

In summarizing, in this paper we consider and optimize DRL-driven non-periodic inspection and maintenance policies in the presence of resource limitations and risk-related constraints. First, the preliminaries of the POMDP formulation in inspection and maintenance planning are elaborated, with insights in the problem-specific modeling requirements. State updating equations and inspection, maintenance, shutdown, and risk cost definitions are presented. It is studied and discussed how the selected actions affect the above costs, and which the inherent mechanisms that drive observational strategies in POMDPs are. Theoretical analysis pertaining to perpetual and instantaneous risk definitions is presented, along with their relation to classical definitions. The optimization problem is cast within the context of decentralized multi-agent DRL control, where agents operate directly on the belief space, i.e. the space of posterior system statistics based on past actions and observations. The developed and employed DRL approach, DDMAC, is an off-policy actor-critic method with experience replay, modifying the original architecture presented in [29]. The relevant algorithmic steps for implementing the above described decentralized DRL framework are provided, based on state augmentation for hard constraints and Lagrangian relaxation for soft constraints. Quantitative investigation is conducted based on a stochastically deteriorating multi-component system. Numerical experiments include evaluation of different baseline policies, and different budget and risk constraint scenarios. The resulting evolution of various system metrics pertaining to risk, reliability, inspection, and intervention choices over the system operating life is parametrically studied and discussed based on the learned policies.

## 2. POMDPs in inspection and maintenance planning

### 2.1. The optimization problem

The goal of the decision-maker (agent) in a life-cycle inspection and maintenance optimization problem is to determine an optimal policy $\pi = \pi^*$ that minimizes the total cumulative future operational costs and risks in expectation:

$$\begin{aligned} \pi^* &= \operatorname*{arg\,min}_{\pi \in \Pi_C \subseteq \Pi} \mathbb{E}_{s_{0:T}, o_{0:T}, a_{0:T}} \left[ \sum_{t=0}^{T} \gamma^t c_t \,\bigg|\, a_t \sim \pi(o_{0:t}, a_{0:t-1}), s_0 \sim \mathbf{b}_0 \right] \\ &= \operatorname*{arg\,min}_{\pi \in \Pi_C \subseteq \Pi} V^\pi(\mathbf{b}_0) \end{aligned} \quad (1)$$

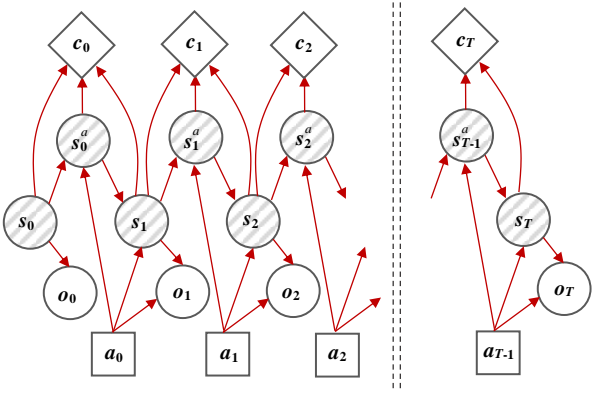

**Fig. 1**. POMDP diagram in time, including intermediate states occurring after actions and before environment transitions.

where $c_t = c(s_t, a_t, s_{t+1})$ is the cost incurred at time $t$ by taking action $a_t \in A$, and transitioning from state $s_t \in S$ to state $s_{t+1} \in S$; $o_t \in \Omega$ is an observation outcome; $\gamma \in [0,1]$ is the discount factor translating future costs to current value; $\mathbf{b}_0$ is an initial distribution over states (or initial belief); $V^\pi$ is the value function, which expresses the total discounted cost given a state or a belief under policy $\pi$; and $T$ is the length of the planning horizon. Planning horizon $T$ can be either finite or infinite. A finite horizon problem can be solved as an infinite one, through proper formulation of the problem, i.e. through augmenting the state space with time, and considering an absorbing state at the final time step [54].

Policy $\pi$ is a rule according to which actions are taken by the decision-maker at different time steps, and it can be, at best, a map from histories of actions and observations to actions, $\pi: A^{t-1} \times \Omega^t \to A$. The policy function belongs to a space, $\pi \in \Pi_c$, which contains all possible policies that are admissible under the existing constraints of the problem. $\Pi_c$ is a subset of $\Pi$, which is the policy space of the unconstrained problem. From the mapping a policy function conducts, it can be observed that the number of possible policies can easily become immense, even in problems with small planning horizons. Also known as the curse of history, this problem is optimally tackled by dynamic programming and POMDPs as explained in detail in the next section. Another approach to attack this complexity, however, often at the expense of solution efficiency, is to exploit problem-specific characteristics and employ simplified assumptions, including approaches that impose action periodicity, policy uniformity among components, component prioritization, ranking, or clustering [12, 55, 56, 57, 58]. Particularly in inspection planning, periodic inspection visits or non-periodic inspections that exploit similarity and/or prioritization of components is typical for deteriorating structural systems [10, 57].

Policy $\pi$ can also be stochastic, in which case it is a mapping to a probability distribution over actions, i.e. $\pi: A^{t-1} \times \Omega^t \to P(A)$. It can be shown under loose regularity conditions about the cost function that the optimal policy in a Markov decision process is deterministic [59]. However, in general and especially in the presence of constraints, the optimal policy is more broadly described by functions accounting for stochastic mappings [42].

### 2.2. Mapping posterior state distributions to actions

In a POMDP environment, transition from state $s_t = s$ to state $s_{t+1} = s'$ is Markovian. Detaching the effect of the maintenance action from the environment transition (natural deterioration), we can define an intermediate state, $s_t^a = s^a \in S$. This state succeeds $s$, with probability $\Pr(s^a|s,a)$, and reflects the system state immediately after maintenance and before the environment transition. This distinction is important to help us better define and quantify the risk in the next section. State $s'$ succeeds $s^a$ with probability $\Pr(s'|s^a,a)$, after the environment transition, i.e. $s' = s^{a,e}$. Owing to the Markovian property, given a pair $(s,a)$, the probability distribution of $s'$ can be fully defined, regardless of the prior history of actions and states as:

$$\Pr(s'|s,a) = \sum_{s^a \in S} \Pr(s'|s^a,a) \Pr(s^a|s,a) \quad (2)$$

Similarly, the cost at a certain time step can be expressed as:

$$c(s,a,s') = \sum_{s^a \in S} \Pr(s^a|s,a) c(s,a,s^a,s') \quad (3)$$

State augmentation can be applied if higher order temporal dependencies exist regarding the history of states and/or actions prior to $t$, or the environment is characterized by non-stationarity [54, 25]. In POMDPs, at each time step, states are hidden to the agent, and are only perceivable through the noisy observation $o_t = o \in \Omega$. Observation $o$ depends on the state of the system and the respective action at the current step, and is defined by probability $\Pr(o|s,a)$. The entire process described above is depicted in the network of Fig. 1.

As a result of the structure of POMDPs, optimal policy $\pi^*$ can be defined, without any loss of information, as a function of belief $\mathbf{b}_t = \mathbf{b} \in B: S \to P(S)$, which is a sufficient statistic of the entire history of previous actions and observations, up to time $t$. Belief $\mathbf{b}$ is thus the posterior probability distribution over states, given the previous belief, and the current transition, action and observation. Hence, the belief at time $t+1$, $\mathbf{b}_{t+1} = \mathbf{b}' = \mathbf{b}^{a,e,o}$, is easily computed though Bayesian updating as:

$$\begin{aligned} b'(s') &= b^{a,e,o}(s') \\ &= \Pr(s'|o',a,\mathbf{b}) \\ &= \frac{\Pr(o'|s',a)}{\Pr(o'|\mathbf{b},a)} b^{a,e}(s') \\ &= \frac{\Pr(o'|s',a)}{\Pr(o'|\mathbf{b},a)} \sum_{s^a \in S} \Pr(s'|s^a,a) b^a(s^a) \end{aligned} \quad (4)$$

where probabilities $b(s)$, for all $s \in S$, form the $|S|$-dimensional belief vector $\mathbf{b}$. The denominator of Eq. (4), $\Pr(o'|\mathbf{b},a)$, is the standard normalizing constant:

$$\Pr(o'|\mathbf{b},a) = \sum_{s' \in S} \Pr(o'|s',a) \sum_{s^a \in S} \Pr(s'|s^a,a) b^a(s^a) \quad (5)$$

Similarly to $s^a$, belief $b^a$ in Eqs. (4) and (5) is the intermediate belief, right after the maintenance action and before the environment transition and observation, defined as:

$$b^a(s^a) = \sum_{s \in S} \Pr(s^a|s,a) b(s) \quad (6)$$

In the special case that the environment is fully observable, i.e. $o=s$, observation specifies exactly which one of the belief vector entries is 1, assigning 0 otherwise. This defines an MDP environment and, accordingly, $\Pr(o|\mathbf{b},a)$ reduces to $\Pr(s'|\mathbf{b},a)$, which is the transition probability of MDPs given the current state distribution. Following this remark, it is apparent that $\Pr(o|\mathbf{b},a)$ holds transition probability semantics for the belief space, $B$, hence a POMDP can be seen as a belief-MDP, where now, however, states are the belief vectors. Accordingly, the transition between beliefs is given as:

$$\Pr(\mathbf{b}'=\mathbf{x}|\mathbf{b},a) = \sum_{o'\in\Omega} \delta_{\mathbf{b}'\mathbf{x}} \Pr(o'|\mathbf{b},a) \quad (7)$$

where $\delta_{ij}$ is the Kronecker delta, i.e. $\delta_{ij}=1$ if $i=j$, 0 otherwise.

This allows us to write the optimality equation, also known as the Bellman equation [22], in the belief space as:

$$\begin{aligned} V(\mathbf{b}) &= HV(\mathbf{b}) \\ &= \min_{a\in A}\{Q(\mathbf{b},a)\} \\ &= \min_{a\in A}\left\{c_b + \gamma \sum_{o'\in\Omega} \Pr(o'|\mathbf{b},a)V(\mathbf{b})\right\} \end{aligned} \quad (8)$$

where $V(\mathbf{b})=V^{\pi^*}(\mathbf{b})$ is the optimal *value function*, representing the total life-cycle cost under the optimal policy $\pi^*$ given an initial belief $\mathbf{b}$, $H$ is the Bellman backup operator, $Q$ is the optimal *action-value function*, and $c_b$ is the expected cost at belief $\mathbf{b}$, defined as:

$$\begin{aligned} c_b &= c_b(\mathbf{b},a) \\ &= \mathbb{E}_{s,s^a,s'}\left[c(s,a,s^a,s')\right] \\ &= \sum_{s\in S} b(s) \sum_{s^a\in S} \Pr(s^a|s,a) \sum_{s'\in S} \Pr(s'|s^a,a) c(s,a,s^a,s') \end{aligned} \quad (9)$$

Operator $H$ is a contraction with unique fixed point $V(\mathbf{b})$. It has been shown that the POMDP cost value function described by the Bellman equation in Eq. (8) is piece-wise linear and concave (convex for the maximization problem) at every time step, composed of linear hyperplanes, also called the *alpha-vectors* [60]. Each alpha-vector corresponds to an inspection and maintenance action [26, 61].

Despite its analogies with MDPs, Eq. (8) is hard to solve exactly through standard MDP-based approaches, e.g. through value or policy iteration. However, there are numerous efficient approximate solution procedures along the lines of *point-based* algorithms [33]. Point-based algorithms sample a subset of the reachable belief space, starting from an initial root belief, thus making value iteration scale linearly with the cardinality of this subset. DRL is used for solving Eq. (8) in this work, using the point-based belief MDP concept combined with deep function approximations and actor-critic training [29].

### 2.3. Risks and costs

Cost at different time steps for a selected action can be decomposed into inspection cost, $c_I$, maintenance cost, $c_M$, and damage state cost, $c_D$. In addition, it is often important for the decision-maker to account for the possibility of additional losses due to intentional system shutdowns, $c_S$, which may occur not as a consequence of damage, but rather as a result of the selected actions. Accounting for this as well, the total cost at each decision step can be generally expressed as:

$$c(s,a,s^a,s') = \underbrace{c_M(s,a)}_{\text{maintenan. cost}} + \underbrace{c_S(s,a)}_{\text{shutdown cost}} + \gamma \underbrace{c_I(s',a)}_{\text{inspection cost}} + \gamma \underbrace{c_D(s^a,s')}_{\text{damage state cost}} \quad (10)$$

Using Eq. (10) in Eq. (9), the expected inspection, maintenance and shutdown costs, can be written as:

$$\begin{aligned} c_{b,X} &= \mathbb{E}_s\left[c_X(s,a)\right] = \sum_{s\in S} b(s) c_X(s,a), \quad X\in\{M,S\} \\ c_{b,I} &= \mathbb{E}_{s'}\left[c_I(a,s')\right] = \sum_{s'\in S} b^{a,e}(s') c_I(a,s') \end{aligned} \quad (11)$$

Although Eq. (11) provides a broad description of the cost function, it is often appropriate to adopt the hypothesis that inspection and maintenance actions affect the respective costs independently, and are also independent of the system state (this hypothesis is stronger for inspections since certain maintenance actions may depend on the extent of damage in the system):

$$\begin{aligned} c_I(s',a) &= c_{b,I}(a) = c_I(a_I) \\ c_M(s,a) &= c_{b,M}(a) = c_M(a_M) \end{aligned} \quad (12)$$

where $a_I \in A_I$ is the selected inspection action and $a_M \in A_M$ is the selected maintenance action. Under this distinctive consideration of actions, the total action can be defined as $a\in A=A_I\times A_M$. We will refer here to no inspection and no maintenance actions as *trivial inspection* and *trivial maintenance actions* respectively. Trivial actions may also refer to routine maintenance and inspection actions, which are actions that are always taken at every time step, thus their costs do not affect the optimization process. Similarly to Eq. (12), it is also reasonable to assume in many problems of inspection and maintenance planning that scheduled shutdowns will be primarily triggered by maintenance actions only, namely:

$$c_S(s,a) = c_S(s,a_M) \quad (13)$$

Damage state cost $c_D$ translates various losses associated with the damage states of the system to cost units. These can be devised into two types of losses, which we will refer to as *instantaneous losses* and *perpetual losses*. Instantaneous losses refer to costs incurred upon arrival at a damage state and do not continue to be collected for as long as the system sojourns this damage state. In the case of a failed civil engineering structure, for instance, such costs can be related to capital-related losses, which occur at the time step at which the structural system transitions to the *failure* state. This cost is collected once over the operating life, unless the system is restored and fails again. Perpetual losses, on the other hand, refer to costs collected for as long as the system sojourns a certain damage state, regardless of which damage state it transitioned from. In the previously mentioned example of a failed civil engineering structure, such costs can be related to economic losses due to downtime, which are, of course, not instantaneous but accrue over time, until the system is restored to an operating status. Following this distinction, the damage cost component of Eq. (10) is written as:

$$c_D(s^a,s') = c_D^{per}(s') + d_{s^a s'} c_D^{inst}(s') \quad (14)$$

where $[d_{ij}]_{i,j \in S}$ is the adjacency matrix pertaining to damage states, as this can be derived by state connectivity according to available actions. That is, if there is an action such that state $j$ is an immediate successor of $i$, then $d_{ij}=1$. For $i=j$, $d_{ij}=0$. In deteriorating environments, it commonly happens that states are ordered, that is, transitions from $s^a$ to $s'$ form an upper-triangular transition matrix, meaning that the system can only transition to a worse state, or at best remain at the same one, due to environment effects. In this case, the adjacency matrix will be strictly upper-triangular.

As implied by Eq. (14), the cost of perpetual losses is a function of the next state, $s'$, whereas the part instantaneous losses depends on the current state after the effect of the maintenance action, $s^a$, and the next state. The expected costs in Eq. (14), which is required to solve Eq. (8), with the aid of Eq. (9), give the step or interval risk as:

$$\begin{aligned} c_{b,D} &= c_{b,D}^{per} + c_{b,D}^{inst} \\ &= \mathbb{E}_{s'}\left[c_D^{per}(s')\right] + \mathbb{E}_{s^a,s'}\left[d_{s^a s'} c_D^{inst}(s')\right] \\ &= \sum_{s^a \in S} b^a(s^a) \sum_{s' \in S} \Pr(s'|s^a, a) \left(c_D^{per}(s') + d_{s^a s'} c_D^{inst}(s')\right) \end{aligned} \quad (15)$$

Using Eq. (15), risk is defined as the expected cumulative discounted damage state cost over the life-cycle:

$$\mathfrak{R}^\pi = \mathbb{E}_{o_{0:T}}\left[\sum_{t=0}^{T} \gamma^t \mathbb{E}_{s_t^a, s_{t+1}}\left[c_D^{per}(s_{t+1}) + d_{s_t^a s_{t+1}} c_D^{inst}(s_{t+1})\right]\right] \quad (16)$$

To better understand Eq. (16), one can consider a case where the system may suffer only instantaneous losses due to failure with cost $c_F$. In this case, Eq. (16) reduces to:

$$\mathfrak{R}_F^\pi = c_F \mathbb{E}_{o_{0:T}}\left[\sum_{t=0}^{T} \gamma^t \left(P_{F_{t+1}|a_{0:t},o_{0:t}} - P_{F_t|a_{0:t},o_{0:t}}\right)\right] \quad (17)$$

where $P_{F_t}$ is the probability of failure up to time $t$. The specialized definition of risk provided by Eq. (17) follows standard risk and reliability assumptions and is well-studied in inspection and maintenance planning [10]. The proof that Eq. (16) reduces to Eq. (17) under the above stated assumptions is presented in Appendix A. This work employs the risk definition of Eq. (16) instead of that of Eq. (17), as it facilitates a broader consideration of losses related to multiple system states.

Similarly, the other step costs of Eq. (10) assume the following expected cumulative discounted values over the life-cycle:

$$\begin{aligned} C_X^\pi &= \mathbb{E}_{o_{0:T}}\left[\sum_{t=0}^{T} \gamma^t \mathbb{E}_{s_t}\left[c_X(s_t, a_t)\right]\right], \; X \in \{M, S\} \\ C_I^\pi &= \mathbb{E}_{o_{0:T}}\left[\sum_{t=0}^{T} \gamma^t \mathbb{E}_{s_{t+1}}\left[c_I(a_t, s_{t+1})\right]\right] \end{aligned} \quad (18)$$

Hence, the optimal POMDP value with its optimality equation described in Eq. (8) is:

$$\begin{aligned} V(\mathbf{b}) &= \min_{\pi \in \Pi_C \subseteq \Pi}\{V^\pi(\mathbf{b})\} \\ &= \min_{\pi \in \Pi_C \subseteq \Pi}\{C_M^\pi + C_S^\pi + \gamma C_I^\pi + \gamma \mathfrak{R}^\pi\} \end{aligned} \quad (19)$$

Thus, overall, the problem of Eq. (1) consists in jointly minimizing the above life-cycle cumulative discounted costs.

### 2.4. To observe or not? Value of information in POMDPs

We can define the step-wise Value of Information (VoI) associated with a certain policy and a certain inspection action $a_I$ as [62]:

$$\text{VoI}_{step}^\pi(a_I) = \mathbb{E}_{o_e}\left[V^\pi\left(\mathbf{b}^{a_M, e, o_e}\right)\right] - \mathbb{E}_{o_e, o_I}\left[V^\pi\left(\mathbf{b}^{a_M, a_I, e, o_e, o_I}\right)\right] \quad (20)$$

Observation $o_e \in \Omega_e$ describes the default observation, i.e. an observation always available to the decision-maker, $o_I \in \Omega_I$ refers to the optional observation provided by the selected inspection action, and $o \in \Omega = \Omega_e \times \Omega_I$ is the total observation.

Similarly, we can define the net step-wise VoI under a policy as:

$$\text{netVoI}_{step}^\pi(a_I) = \text{VoI}_{step}^\pi - c_{b,I} \quad (21)$$

Net step-wise VoI expresses the net gain as a result of additional information, also considering the cost to acquire this information (i.e. inspection cost). Elaborating on Eq. (8), and considering the fact that inspections do not change the state of the system, we have [62]:

$$V(\mathbf{b}) = \min_{a_M \in A_M}\left\{c_{b,I^-} + \gamma \mathbb{E}_{o_e}\left[V\left(\mathbf{b}^{a_M, e, o_e}\right)\right] - \gamma \max_{a_I \in A_I}\left\{\text{netVoI}_{step}^{\pi^*}(a_I)\right\}\right\} \quad (22)$$

where $c_{b,I^-}$ combines any costs other than the expected inspection cost, i.e. maintenance, shutdown, and damage state costs. Eq. (22) provides an alternative description of the Bellman equation, and shows that for any possible maintenance action, the decision-maker takes that inspection action which maximizes the net VoI at this step.

Following the above, the concavity of the POMDP value function of Eq. (8), and the properties of the Bellman contraction operator, we can show that step-wise VoI, as well as VoI over the life-cycle, are always non-negative if the decisions follow the POMDP optimal policy [62, 63]. At the extreme case that no inspection means no information at all, VoI reaches its highest value. This result can be similarly shown.

### 3. Operating under constraints

We consider the following form of the stochastic optimization problem of Eq. (1):

$$\pi^* = \arg\min_{\pi \in \Pi} \mathbb{E}_{s_{0:T}, o_{0:T}, a_{0:T}}\left[\sum_{t=0}^{T} \gamma^t c_t \, \bigg| \, a_t \sim \pi(o_{0:t}, a_{0:t-1}), s_0 \sim \mathbf{b}_0\right] \quad (23)$$

$$\text{s.t. } G_{h,k} = \sum_{t=0}^{T} \gamma^t g_{h,k}(s_t, a_t) - \alpha_{h,k} \leq 0, \; k=1,..,K$$

$$G_{s,m} = \mathbb{E}_{s_{0:T}, o_{0:T}, a_{0:T}}\left[\sum_{t=0}^{T} \gamma^t g_{s,m}(s_t, a_t, s_{t+1})\right] - \alpha_{s,m} \leq 0, \; m=1,..,M$$

where $G_{h,k}$ and $G_{s,m}$ are the *hard* and *soft* constraints, respectively, $g_{h,k}$ and $g_{s,m}$ are their respective auxiliary costs (e.g. $c_M$, $c_I$, $c_S$, $c_D$, or else), and $\alpha_{h,k}$, $\alpha_{s,m}$ are real-valued scalars. The form of constraints in Eq. (23) is amenable to a broad class of constraint types that are relevant to infrastructure management. For example, hard constraints can model a great variety of fixed resource allocation and control action availability problems, such as problems referring to strict budget limitations. In turn, soft constraints, of the Eq. (23) form, can model a great variety of risk-based constraints. More details about these can be found in Section 3.2. The term *soft* constraints, although not standard in stochastic optimization and optimal control literature, is used here to distinguish from the term *hard* constraints, indicating that the underlying constraints do not need to be strictly satisfied, but are rather imposed in an expected or probabilistic fashion.

Hard constraints can be straightforwardly taken into account through state augmentation. On an interesting remark, in one of his notes on dynamic programming under constraints in 1956 [64], R. Bellman mentions that this approach may not be favored since "due to the limited memory of present-day digital computers, this method founders on the reef of dimensionality". However, this restriction has been widely lifted today, whereas DRL has diminished the effects of the curse of state dimensionality even further. Thus, state augmentation is followed for the hard constraints here. Note that, in the special case where functions $g_{s,m}$ are deterministic, soft constraints become hard. However, soft constraints are not practical to consider through state augmentation since one should track the entire distribution of the cumulative discounted value of $g_{s,m}$. Therefore, probabilistic constraints are addressed here through Lagrangian relaxation [65]. Based on the above, the optimization problem is restated as:

$$V(\mathbf{b}) = \max_{\lambda_1,...,\lambda_M \geq 0} \min_{\pi \in \Pi} \mathbb{E}_{s_{0:T}, y_{0:T}, o_{0:T}, a_{0:T}} \left[ \sum_{t=0}^{T} \gamma^t \left( c(s_t, a_t, s_{t+1}) \right. \right.$$
$$\left. \left. + \sum_{m=1}^{M} \lambda_m (g_{s,m} - \alpha_{s,m}) \right) \middle| a_t \sim \pi(o_{0:t}, a_{0:t-1}, \mathbf{y}_t), s_0 \sim \mathbf{b}_0, \mathbf{y}_0 \right]$$
$$= \max_{\lambda_1,...,\lambda_M \geq 0} \min_{\pi \in \Pi} V_\lambda^\pi(\mathbf{b}_0, \mathbf{y}_0)$$
$$\text{s.t. } \mathbf{y}_t = \{y_{kt}\}_{k=1}^{K}, y_{kt} = \sum_{\tau=0}^{t-1} \gamma^\tau g_{h,k}(s_\tau, a_\tau), y_{k0} = 0, \quad (24)$$
$$y_{kt} \in [0, \alpha_{h,k}], k = 1,...,K$$

where variables $y_{kt}$ track the discounted cumulative value of the function related to hard constraints, $g_{h,k}$, up to time step $t$-1. Variables $y_{kt}$ are upper bounded by $a_{h,k}$. Lagrange multipliers, $\lambda_m$, constitute the dual variables of the max-min dual problem, they are positive scalars, and are associated with the soft constraints.

### 3.1. Budget constraints

Depending on the operational and resource allocation strategy of the management agency, available funding for inspection and maintenance must comply with certain short-term or long-term goals related to a specific budget cycle duration, $T_B$. Namely, in the extreme case of a short-term budget cycle duration, budget caps exist for every decision step (e.g. annual budget), whereas in the extreme case of a long-term budget cycle duration, there is a budget cap pertaining to the cumulative inspection and maintenance expenses over the entire life-cycle of the system, i.e. $T_B=T$. The cumulative cost of inspection and maintenance actions over period $T_B$ is given for:

$$g_h(s_\tau, a_\tau) = (c_M + \gamma c_I) \mathbf{1}_{\tau \in \Lambda_t} \quad (25)$$

$$\Lambda_t = \left( \lfloor t/T_B \rfloor T_B, (\lfloor t/T_B \rfloor + 1) T_B \right] \quad (26)$$

where $\lfloor x \rfloor$ is the integer part of $x$. For a given budget cap $\alpha_h$, the maintenance and inspection costs at each time step read:

$$\bar{c}_M = \mathbf{1}_{y+g_h \leq \alpha_h} c_M$$
$$\bar{c}_I = \mathbf{1}_{y+g_h \leq \alpha_h} c_I \quad (27)$$

According to Eqs. (25)-(27), inspection and maintenance costs are accounted for only at the current budget cycle, and if the currently selected action does not violate the budget cap. The total cost at each time step of Eq. (10) is accordingly rewritten as:

$$\bar{c}_t = \bar{c}_M(s_t, a_t) + \gamma \bar{c}_I(a_t, s_{t+1}) + c_S(s_t, a_t) + \gamma c_D(s_t^a, s_{t+1}) \quad (28)$$

Transition and observation probabilities are also affected by the presence of the budget constraints as:

$$\Pr(s^a | s, y, a) = \mathbf{1}_{y+g_h \leq \alpha_h} \Pr(s^a | s, a) + (1 - \mathbf{1}_{y+g_h \leq \alpha_h}) \Pr(s^a | s, a_o)$$
$$\Pr(s' | s^a, y, a) = \mathbf{1}_{y+g_h \leq \alpha_h} \Pr(s' | s^a, a) + (1 - \mathbf{1}_{y+g_h \leq \alpha_h}) \Pr(s' | s^a, a_o) \quad (29)$$
$$\Pr(o' | s', y, a) = \mathbf{1}_{y+g_h \leq \alpha_h} \Pr(o' | s', a) + (1 - \mathbf{1}_{y+g_h \leq \alpha_h}) \Pr(o' | s', a_o)$$

where $a_o$ is the trivial decision, where no inspection and no maintenance are performed. As indicated by Eqs. (24)-(29), incorporation of budget constraints can be accomplished by accounting for new state variables $y=y_t$. This way the agent is able to reason about control actions based on the available budget, $\alpha_h$ - $y_t$, at each time step of the decision process. In the case of step-wise budget constraints, i.e. $T_B=1$, this state augmentation is not necessary, since the agent does not need to track any inspection and maintenance expenses made in the past, thus having the entire amount of $\alpha_h$ at its disposal for every single step.

As opposed to state variables $s_t$, new variables $y_t$ are fully observable. In this regard, the problem can be also seen as a mixed observability Markov decision process, which admits favorable state decompositions and can be solved by value iteration algorithms in settings with moderate dimensions [18]. In this case, constrained value iteration based POMDP solution procedures devised for constrained problems can be employed to drive the optimization process [44, 45, 46]. However, as for the unconstrained case, such formulations can manifest limitations related to efficient scaling in systems with large state and action spaces, like the systems that are typically encountered in the class of sequential decision-making for infrastructure and networks.

### 3.2. Risk-based constraints

For notational efficiency of the present section we introduce the following random variables:

$$J_i^\pi = \sum_{t=0}^{T} \gamma^t c_i(s_t, a_t, s_{t+1}), i \in \{M, I, S, D\}$$
$$J^\pi = \sum_i J_i^\pi \quad (30)$$

where $J_M^\pi$, for example, accumulates total costs, related to maintenance actions over the life-cycle, and $\mathbb{E}_{s_{0:T}, o_{0:T}, a_{0:T}}[J_M^\pi] = C_M^\pi$ according to the definitions of Eq. (11).

We are now interested in incorporating constraints that bound risk over the system life-cycle. The risk-related random variable based on Eqs. (16) and (30) is $J_D^\pi$. Thus, the respective constraint function obtains the following form, for $g_s = c_D$ in Eq. (23):

$$\begin{aligned} G_s &= \mathbb{E}_{s_{0:T}, o_{0:T}, a_{0:T}} \left[ J_D^\pi \right] - \alpha_s \\ &= \mathbb{E}_{s_{0:T}, o_{0:T}, a_{0:T}} \left[ \sum_{t=0}^{T} \gamma^t \left( c_D^{per}(s_{t+1}) + d_{s_t^a s_{t+1}} c_D^{inst}(s_{t+1}) \right) \right] - \alpha_s \\ &= \Re^\pi - \alpha_s \end{aligned} \quad (31)$$

It should be noted that, although the budget constraints of focus in this work are not soft, budget constraints can also be expressed through $G_s$ constraints, satisfied in expectation, depending on the modeling needs of the problem, as in Eq. (31). Any other costs as introduced in Eq. (10) can be considered in the same logic as well.

Constraints of the generic $G_s$ form are also the *chance* or *probabilistic* constraints, which bound the probabilities of certain quantities or events [52, 53]. As such, if one wants to bound the probability of the optimal policy exceeding a certain life-cycle cost threshold $J_{cr}$, one may apply the following $g_s$ function, for any $J_i^\pi$ similarly to Eq. (31):

$$g_s = \mathbf{1}_{t=T} \cdot \mathbf{1}_{J_i^\pi > J_{cr}} \quad (32)$$

where the second indicator signifies the cumulative cost constraint violation, and the first one ensures that this is taken into account once, at the end of the planning horizon. Taking the expectation of cumulative value of the constraint function of Eq. (32), we have:

$$\begin{aligned} G_s &= \mathbb{E}_{s_{0:T}, o_{0:T}, a_{0:T}} \left[ \sum_{t=0}^{T} \mathbf{1}_{t=T} \cdot \mathbf{1}_{J_i^\pi > J_{cr}} \right] - \alpha_s \\ &= \Pr\left( J_i^\pi > J_{cr} \right) - \alpha_s \end{aligned} \quad (33)$$

Considering Eq. (33), if $\alpha_s = 1$, we end up with a hard constraint requirement, i.e. $J_i^\pi > J_{cr}$. It is thus obvious that hard constraints can be also seen as a limiting case of soft constraints.

From a stricter reliability standpoint, many decision problems are interested in bounding the probability of failure (i.e. the probability reaching a failure state $s_F$ from a non-failure state) over the system operating life. In this case, we just need to set, $c_D^{per} = 0$, $\gamma = 1$, and $c_D^{inst} = \delta_{s_{t+1} s_F}$ in Eq. (31):

$$\begin{aligned} G_s &= \mathbb{E}_{s_{0:T}, o_{0:T}, a_{0:T}} \left[ \sum_{t=0}^{T} d_{s_t^a s_{t+1}} \delta_{s_{t+1} s_F} \right] - \alpha_s \\ &= P_{F_T} - \alpha_s \end{aligned} \quad (34)$$

$P_{F_T}$ is the probability of failure up to the end of the life-cycle $t=T$. Scalar $\alpha_s$ in Eq. (33) and (34) is a valid probability designating the (1 - $\alpha_s$)-percentile of risk and probability of failure, respectively, the decision-maker is willing to tolerate.

Other relevant constraint definitions in stochastic optimization and constrained Markov decision processes literature include constraints on the *value-at-risk* and *conditional-value-at-risk* [66, 53] (with the former coinciding with probabilistic constraints), constraints on the policy variance [67, 68], as well as constraints whose satisfaction is implicitly encouraged through reward-based penalization [69].

### 3.3. Constrained control with deep reinforcement learning

In recent work by the authors [29, 30], the Deep Centralized Multi-agent Actor Critic approach has been proposed for management of large engineering systems, which treats system control units as individual agents making decentralized decisions based on shared/centralized system information and actor hidden layer parameters. Control units are defined in reference to system parts for which separate actions apply at each decision step, and can be either individual system components or greater sub-system parts comprised of multiple components. As such, one control unit has at least one component, and one component may belong to more than one control units. The system policy function is written as:

$$\pi(\mathbf{a} \mid \hat{\mathbf{b}}, \mathbf{y}) = \prod_{i=1}^{N_{CU}} \pi_i\left( a^{(i)} \mid \hat{\mathbf{b}}, \mathbf{y} \right) \quad (35)$$

where $\mathbf{a}$ is a vector of actions and $\hat{\mathbf{b}}$ is a 2D matrix, such that:

$$\begin{aligned} \mathbf{a} &= \left[ a^{(i)} \right]_{i=1}^{N_{CU}} \\ \hat{\mathbf{b}} &= \left[ \mathbf{b}^{(j)} \right]_{j=1}^{N_C} \end{aligned} \quad (36)$$

where $a^{(i)}$ is the action of control unit $i$, $\mathbf{b}^{(j)}$ is the belief of system component $j$, $N_{CU}$ is the number of control units, and $N_C$ is the number of system components.

The policy functions of Eq. (35), as well as a centralized system Lagrangian value function are parametrized with the aid of deep neural networks as:

$$\begin{aligned} \pi_i\left( a^{(i)} \mid \hat{\mathbf{b}}, \mathbf{y} \right) &\simeq \pi_i\left( a^{(i)} \mid \hat{\mathbf{b}}, \mathbf{y}, \boldsymbol{\theta}_\pi^{(i)} \right) \\ V_\lambda^\pi\left( \hat{\mathbf{b}}, \mathbf{y} \right) &\simeq V_\lambda^\pi\left( \hat{\mathbf{b}}, \mathbf{y} \mid \boldsymbol{\theta}_V \right) \end{aligned} \quad (37)$$

Parameters $\boldsymbol{\theta}_\pi^{(i)}$, $\boldsymbol{\theta}_V$ are real-valued vectors, and can either vary or be shared among control units. In either case, each control unit's policy is conditioned on the global belief and the budget-related states. Note that here we have a separate policy network for each agent as denoted by superscript $i$ in the policy parameters of Eq. (37), thus a completely decentralized actor parametrization is used. To distinguish this from the original DCMAC architecture we call this Deep Decentralized Multi-agent Actor Critic (DDMAC). As discussed in Section 1, both provide decentralized POMDP policy

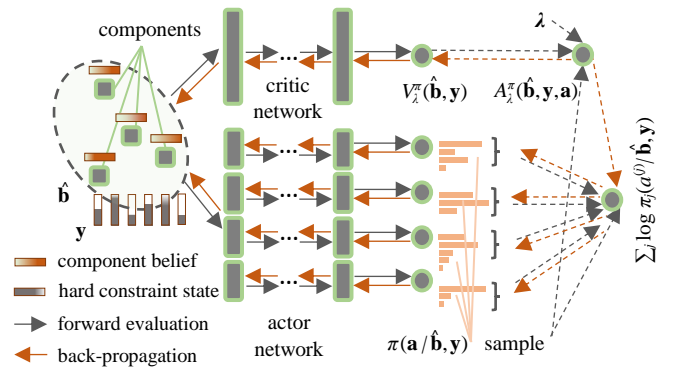

**Fig. 2.** Constrained Deep Decentralized Multi-agent Actor Critic (DDMAC) architecture.

**Algorithm 1** Constrained Deep Decentralized Multi-agent Actor Critic

Initialize replay buffer
Initialize actor, critic, and dual parameters $\left[\boldsymbol{\theta}_{\pi}^{(j)}\right]_{j=1}^{N_{CU}}, \boldsymbol{\theta}_V, [\lambda_m]_{m=1}^M$
**for** *number of episodes* **do**
    **for** *t=1,…,T* **do**
        Select action $\mathbf{a}_t$ at random according to exploration noise
        Otherwise select action $\mathbf{a}_t \sim \boldsymbol{\mu}_t = \left[\pi_j\left(\cdot \mid \hat{\mathbf{b}}_t, \mathbf{y}_t, \boldsymbol{\theta}_\pi^{(j)}\right)\right]_{j}^{N_{CU}}$
        Estimate costs $\bar{c}_{b,t} = \bar{c}_b$, $g_{s,mt} = g_{s,m}$ given $\hat{\mathbf{b}}_t$ and $\mathbf{a}_t$
        Observe $o_{t+1}^{(l)} \sim p\left(o_{t+1}^{(l)} \mid \mathbf{b}_t^{(l)}, \mathbf{y}_t, \mathbf{a}_t\right)$ for $l=1,2,...,N_C$
        Compute beliefs $\mathbf{b}_{t+1}^{(l)}$ for $l=1,2,...,N_C$
        Store tuple $\left(\hat{\mathbf{b}}_t, \mathbf{y}_t, \mathbf{a}_t, \boldsymbol{\mu}_t, \bar{c}_{b,t}, [g_{s,mt}]_{m=1}^M, \hat{\mathbf{b}}_{t+1}, \mathbf{y}_{t+1}\right)$ to replay buffer
        Sample batch $\left(\hat{\mathbf{b}}_i, \mathbf{y}_i, \mathbf{a}_i, \boldsymbol{\mu}_i, \bar{c}_{b,i}, [g_{s,mi}]_{m=1}^M, \hat{\mathbf{b}}_{i+1}, \mathbf{y}_{i+1}\right)$ from replay buffer
        If $\hat{\mathbf{b}}_i$ is terminal state $A_{\lambda,i}^\pi = \bar{c}_{b,i} + \sum_{m=1}^M \lambda_m g_{s,mi} - V_\lambda^\pi\left(\hat{\mathbf{b}}_i, \mathbf{y}_i \mid \boldsymbol{\theta}^V\right)$
        Otherwise $A_{\lambda,i}^\pi = \bar{c}_{b,i} + \sum_{m=1}^M \lambda_m g_{s,mi} + \gamma V_\lambda^\pi\left(\hat{\mathbf{b}}_{i+1}, \mathbf{y}_{i+1} \mid \boldsymbol{\theta}^V\right) - V_\lambda^\pi\left(\hat{\mathbf{b}}_i, \mathbf{y}_i \mid \boldsymbol{\theta}^V\right)$
        Update actor parameters $\boldsymbol{\theta}_\pi^{(j)}$ according to gradient:
$$\nabla_{\boldsymbol{\theta}_\pi^{(j)}} V_\lambda^\pi \simeq \sum_i w_i \left(\sum_{j=1}^{N_{CU}} \nabla_{\boldsymbol{\theta}^\pi} \log \pi_j\left(a_i^{(j)} \mid \hat{\mathbf{b}}_i, \mathbf{y}_i, \boldsymbol{\theta}_\pi^{(j)}\right)\right) A_{\lambda,i}^\pi$$
        Update critic parameters $\boldsymbol{\theta}^V$ according to gradient:
$$\nabla_{\boldsymbol{\theta}_V} V_\lambda^\pi \simeq \sum_i w_i \nabla_{\boldsymbol{\theta}^V} V_\lambda^\pi\left(\hat{\mathbf{b}}_i, \mathbf{y}_i \mid \boldsymbol{\theta}^V\right) A_{\lambda,i}^\pi$$
        Update dual variables $\lambda_m$, $m=1,…,M$, based on current policy return, according to gradient:
$$\nabla_{\lambda_m} V_\lambda^\pi \simeq \sum_{t=0}^T \gamma^t g_{s,mt} - \alpha_{s,m}$$
    **end for**
**end for**

solutions. The respective architectures are shown in Fig. 2. In this figure, 4 components are depicted, and each component is a control unit, thus $N_{CU}=N_C$.

DDMAC is trained based on off-policy experiences retrieved from the replay memory or replay buffer, $\mathcal{M}$. These experiences are in the form of $\left(\hat{\mathbf{b}}, \mathbf{y}, \mathbf{a}, [\pi_i]_{i=1}^{N_{CU}}, \bar{c}_b, [g_{s,m}]_{m=1}^M, \hat{\mathbf{b}}', \mathbf{y}'\right)$ tuples that are generated while the agent interacts with the environment. Thus, the replay memory is a stack of transition tuples.

The off-policy gradients of the policy function and the value function are computed by importance sampling as:

$$\nabla_{\boldsymbol{\theta}_\pi^{(i)}} V_\lambda^\pi = \mathbb{E}_\mathcal{M}\left[w_t \left(\sum_{i=1}^{N_{CU}} \nabla_{\boldsymbol{\theta}^\pi} \log \pi_i\left(a^{(i)} \mid \hat{\mathbf{b}}, \mathbf{y}, \boldsymbol{\theta}_\pi^{(i)}\right)\right) A_\lambda^\pi\left(\hat{\mathbf{b}}, \mathbf{y}, \mathbf{a}\right)\right] \quad (38)$$

$$\nabla_{\boldsymbol{\theta}_V} V_\lambda^\pi = \mathbb{E}_\mathcal{M}\left[w_t \nabla_{\boldsymbol{\theta}^V} V_\lambda^\pi\left(\hat{\mathbf{b}}, \mathbf{y} \mid \boldsymbol{\theta}^V\right) A_\lambda^\pi\left(\hat{\mathbf{b}}, \mathbf{y}, \mathbf{a}\right)\right] \quad (39)$$

where $w_t$ is the importance sampling weight with sample distribution a policy $\boldsymbol{\mu}$ retrieved from the experience replay and target distribution the current policy. $A_\lambda^\pi$ is the advantage function, which is herein approximated by the temporal difference:

$$A_\lambda^\pi\left(\hat{\mathbf{b}}, \mathbf{y}, \mathbf{a} \mid \boldsymbol{\theta}^V\right) \simeq \bar{c}_b + \sum_{m=1}^M \lambda_m g_{s,m}$$
$$+ \gamma V_\lambda^\pi\left(\hat{\mathbf{b}}', \mathbf{y}' \mid \boldsymbol{\theta}^V\right) - V_\lambda^\pi\left(\hat{\mathbf{b}}, \mathbf{y} \mid \boldsymbol{\theta}^V\right) \quad (40)$$

The gradient of dual variables $\lambda_m$ is easily computed as [50]:

$$\nabla_{\lambda_m} V_\lambda^\pi \simeq \sum_{t=0}^T \gamma^t g_{s,m} - \alpha_{s,m} \quad (41)$$

Dual variables are updated through on-policy samples since off-policy weighted sampling of multiple time steps produces high-variance estimators that may trigger training instabilities. Algorithm 1 describes the aforementioned implementation steps.

## 4. Results

### 4.1. Environment details

A stochastic, non-stationary, partially observable 10-component deteriorating system is considered, operating over a life-cycle period of 50 decision steps (years), with a discount factor of $\gamma=0.975$. For civil engineering systems, discount factors typically range from 0.93 to 0.98. Higher discount factors make the decision problem more challenging, in the sense that they increase the effective horizon of important decisions. Links between components create the system shown in Fig. 3. It is assumed that link operation depends solely on the operating status of the respective components. Overall system connectivity is determined by the connectivity of nodes A and B.

Each component has independent deterioration dynamics. These are expressed by 4x4x50 three-dimensional transition matrices, corresponding to 4 damage states (*intact*, *minor damage*, *major damage*, *severe damage*), combined with 50 deterioration rates, as many as the decision steps of the system life-cycle. Component transitions are given in Tables 1,2. Component transition parameters for the underlying hidden Markov models are assumed to be known or already learned, thus model uncertainty is not considered in this example. For learning of hidden Markov models based on structural data the interested reader is referred to [70, 71]. Different failure probabilities are considered based on each one of the above damage states, as shown in Table 3. Thus, the system behavior, as a whole, is described by the Bayesian network of Fig. 4.

Further details on consistently coupling inference of dynamic Bayesian networks with POMDPs for deteriorating structures can be found in [72, 73]. The final state vector for each component is $s^{(i)}=(x^{(i)}, \tau^{(i)}, f^{(i)}, t)$, where $x^{(i)}$ is the damage state; $\tau^{(i)}$ is the deterioration rate; $f^{(i)}$ is a binary failure indicator; and $t$ is the decision time step ($t$ is the same for all components). Failure is considered an absorbing state. Hence, we assume that when a component fails it remains failed at the next step, as long as no restorative action is taken. This allows us to augment the component state space, finally obtaining 5x5x50 transition matrices.

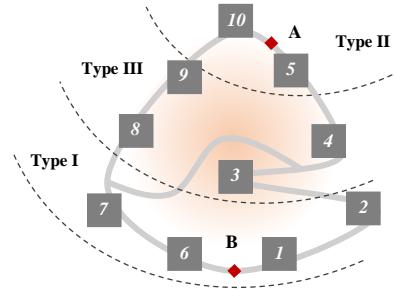

**Fig. 3**. Multi-component deteriorating system. System fails when connectivity between nodes A and B is lost. Major costs are incurred when system fails. Minor costs are incurred for combinations of failed series subsystems. Types I-III refer to the severity of the deterioration model, from less to more severe, respectively.

**Table 1**
Component initial damage state transition probabilities for deterioration model Types I, II, and III.

| Deterioration model | $p_{12}$ | $p_{13}$ | $p_{14}$ | $p_{23}$ | $p_{24}$ | $p_{34}$ |
|---|---|---|---|---|---|---|
| Type I | 0.0129 | 0.0072 | 0.0008 | 0.0102 | 0.0038 | 0.0092 |
| Type II | 0.0311 | 0.0096 | 0.0014 | 0.0283 | 0.0057 | 0.0281 |
| Type III | 0.0428 | 0.0229 | 0.0033 | 0.0406 | 0.0095 | 0.0328 |

**Table 2**
Component final damage state transition probabilities for deterioration model Types I, II, and III.

| Transition probability | $p_{12}$ | $p_{13}$ | $p_{14}$ | $p_{23}$ | $p_{24}$ | $p_{34}$ |
|---|---|---|---|---|---|---|
| Type I | 0.0618 | 0.0512 | 0.0036 | 0.0905 | 0.0091 | 0.0768 |
| Type II | 0.0862 | 0.0868 | 0.0051 | 0.1219 | 0.0121 | 0.1091 |
| Type III | 0.1347 | 0.0669 | 0.0098 | 0.1665 | 0.0244 | 0.1462 |

**Table 3**
Component failure probabilities for different deterioration types and damage states.

| Damage state | intact | minor | major | severe |
|---|---|---|---|---|
| Type I | 0.0019 | 0.0067 | 0.0115 | 0.0177 |
| Type II | 0.0028 | 0.0076 | 0.0163 | 0.0219 |
| Type III | 0.0088 | 0.0210 | 0.0449 | 0.0564 |

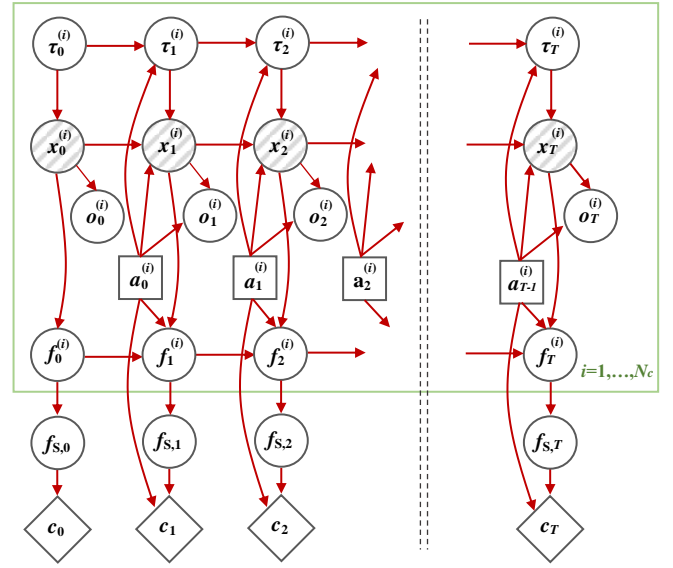

**Fig. 4**. Dynamic Bayesian network of multi-component deteriorating system in time.

We consider three types of available maintenance actions; $A_M$ ={*no-repair*, *partial-repair*, *restoration/replacement*}. There are also two types of available inspection actions; $A_I$ ={*no-inspection*, *inspection*}. Accordingly, to allow for utmost diversification between component policies, each component, which herein defines a separate control unit, is assigned 5 available inspection and maintenance actions, based on the combinations of the abovementioned sets, i.e. $a^{(i)} \in A_M \times A_I \setminus (\textit{restoration/replacement}, \textit{inspection})$. The (*restoration/replacement*, *inspection*) action is excluded from the set of available actions, as it is assumed that whenever a system component is replaced, thus returning to an as good as new condition, a decision for inspection is strictly suboptimal. No-repair costs are null, whereas restoration/replacement costs are the same for all components. Partial-repair costs are (7.5,10,15)% of the component replacement cost, for component Types (I, III, II), respectively. Inspection costs are the same for all components, at 1.5% of the component replacement cost. Partial-repairs send components one damage state back without changing the deterioration rate, restorations/replacements send components to the initial damage state and deterioration rate, whereas no-repairs have no effect on the damage state and deterioration rate. Partial-repairs have no effect on failed components and are considered to have been completed before the next environment transition. When restorations/replacements are chosen, these are completed at the end of the next time step, negating the deterioration transition during that step. Thus, in this case, the next state is the intact one with certainty.

If an inspection action is taken, observation probabilities are given by the following observation matrices:

$$\left[ \Pr\left(o^{(i)} \mid s^{(i)}, a^{(i)} \in A_M \times \{inspection\}\right) \right]_{\substack{o^{(i)} \in \Omega \\ s^{(i)} \in S}} =$$

$$= \begin{bmatrix} 0.84 & 0.13 & 0.02 & 0.01 & \\ 0.11 & 0.77 & 0.09 & 0.03 & \\ 0.02 & 0.16 & 0.70 & 0.12 & \\ 0.01 & 0.02 & 0.13 & 0.84 & \\ & & & & 1 \end{bmatrix} \quad (42)$$

Observation matrices depend on state discretization and presumed measurement noise or estimated model errors [15]. Failure is considered to be a self-announcing event, hence, component (5,5) of the observation matrix of Eq. (42) is 1. Accordingly, if no inspection is taken the observation matrix reads:

$$\left[ \Pr\left(o^{(i)} \mid s^{(i)}, a^{(i)} \in A_M \times \{no-inspection\}\right) \right]_{\substack{o^{(i)} \in \Omega \\ s^{(i)} \in S}} =$$

$$= \begin{bmatrix} 1 & 1 & 1 & 1 & \\ & & & & 1 \end{bmatrix}^T \quad (43)$$

System failure, i.e. loss of connectivity between nodes A and B, is described by random variable $f_s$. Random variable $f_s$ assumes 4 values associated with events $E_0$: all links available, $E_1$: 25% of links failed, $E_2$: 50% of links failed without system failure, and $F_s$: system failure. A link is failed if at least one component is failed. We can thus consider the series subsystems, controlling the link failures, $l_1$={1,2,3}, $l_2$={4,5}, $l_3$={6,7} and $l_4$={8,9,10}. Their failure events are accordingly described by events $F_{l,1}$, $F_{l,2}$, $F_{l,3}$, and $F_{l,4}$. Based on the above, it can be derived that the system failure probability is:

$$\Pr(F_s) = \Pr(F_{l,1})\Pr(F_{l,3}) + \Pr(F_{l,2})\Pr(F_{l,4}) - \prod_{i=1}^{4} \Pr(F_{l,i}) \quad (44)$$

The corresponding non-failure events of interest, $E_0$, $E_1$, $E_2$, are defined as:

$$E_0 : \bigcap_{i=1}^{4} F_{l,i}^-$$
$$E_1 : \bigcup_{i=1}^{4} \left( F_{l,i} \bigcap_{j \neq i}^{4} F_{l,j}^- \right) \quad (45)$$
$$E_2 : \left( \bigcup_{i,j=1, i>j}^{4} F_{l,i} \cap F_{l,j} \right) \cap F_s^-$$

Accordingly, the probabilities of events $E_1$, $E_2$ are computed as:

$$\Pr(E_0) = \prod_{i=1}^{4} \left(1 - \Pr(F_{l,i})\right)$$
$$\Pr(E_1) = \sum_{i=1}^{4} \prod_{j=1, j \neq i}^{4} \left(1 - \Pr(F_{l,j})\right) \Pr(F_{l,i}) \quad (46)$$
$$\Pr(E_2) = 1 - \Pr(E_0) - \Pr(E_1) - \Pr(F_s)$$

Perpetual and instantaneous losses due to failure are equivalent to 2.5 and 50 times the system rebuild cost, respectively, i.e. $c_{F_s}^{per} = 2.5 \cdot c_{reb}$ and $c_{F_s}^{inst} = 50 \cdot c_{reb}$, respectively. Similarly, we consider perpetual and instantaneous losses incurred when 25% and 50% of system links are not available (i.e. at least one of their respective components is at the failure state). These losses are incurred if events $E_1$, $E_2$ occur, respectively, and are quantified in cost units as $c_{E_1}^{per} = 0.05 \cdot c_{reb}$, $c_{E_2}^{per} = 0.25 \cdot c_{reb}$, $c_{E_1}^{inst} = 1 \cdot c_{reb}$, $c_{E_2}^{inst} = 5 \cdot c_{reb}$. In case of transportation networks, for example, such perpetual losses may refer to time delays and/or additional user costs due to detours, whereas such instantaneous losses may pertain to capital loss due to asset failures related to those links.

Based on the above losses, the fact that system events are fully observable, and following the risk definition of Eq. (16), the interval risk reads:

$$c_{b,D} = \sum_{\substack{f_s \in \\ \{F_s, E_2, E_1\}}} \Pr(f_{s,t+1}) \left( c_{f_{s,t+1}}^{per} + \left(1 - \Pr(f_{s,t})\right) c_{f_{s,t+1}}^{inst} \right) \quad (47)$$

Apart from the above losses, additional costs are included in the analysis, pertaining to scheduled system shutdowns. Those come as a result of different action combinations on different system components. That is, considering that non-trivial maintenance actions require some degree of component non-operability for completion during a time step, events $E_{a0}$, $E_{a1}$, $E_{a2}$, $F_{as}$ can occur, in analogy to events $E_0$, $E_1$, $E_2$, $F_s$. Those losses are only incurred if the system would be otherwise in an operating condition, i.e. not failed. Events and their probabilities are similarly defined as in Eqs. (45)-(47), whereas respective costs are the same as the perpetual losses due to events $E_0$, $E_1$, $E_2$, $F_s$.

### 4.2. Experimental setup

For the purposes of this numerical investigation two sets of analyses a conducted. The first set considers a budget cycle period of $T_B = 5$. Each budget period shares the same budget cap, and 9 different levels of budget constraints are considered, which are given as functions of the system rebuild cost, {5, 7.5, 10, 12.5, 15, 17.5, 20, 25, 30}% $c_{reb}$. For the second set of analyses, 9 different levels of life-cycle risk constraints are considered, i.e. {1, 1.25, 1.5, 1.75, 2, 2.25, 2.5, 2.75, 3.25} $c_{reb}$. In addition to the above analyses, the unconstrained policy is also learned.

For training, the Keras deep learning python libraries are used with Tensorflow backend. For all analyses, the actor networks consist of two fully connected hidden layers with 50 Rectified Linear Unit activation functions each, for all 10 components. No parameters are shared among component actors, and each control unit has a 5-dimensional softmax output corresponding to the cardinality of $A_M \times A_I \setminus$ (*restoration/replacement*, *inspection*). The critic network also consists of two fully connected hidden layers with 150 ReLU activations each. The critic has a one-dimensional linear output, which approximates the POMDP Lagrangian value function of the entire system.

The Adam optimizer [74] is utilized for stochastic gradient descent on the networks parameter space, with learning rates being gradually adjusted from 1E-3 and 1E-4, to 1E-4 and 1E-5 for the critic and actor, respectively. The learning rate of Lagrange multipliers is set to 1E-5. The size of the experience replay is set equal to 300,000 and an exploration noise of linearly annealed from 100% to 1% is added at the first 2500 episodes of the training process.

### 4.3. DRL solutions and baseline policies

To verify the quality of DDMAC solutions, we construct and optimize various baseline policies, incorporating well-established condition- risk-, and time-based inspection and maintenance assumptions, which are also combined with periodic action considerations, as well as component prioritization approaches. These baselines are:

- Fail Replacement (FR) policy. No inspections are taken. If a component fails it is immediately replaced. No variable is optimized.
- Age-Periodic Maintenance (APM) policy. No inspections are taken, whereas maintenance actions are taken based on the age of components. Two maintenance ages are optimized; periodic age

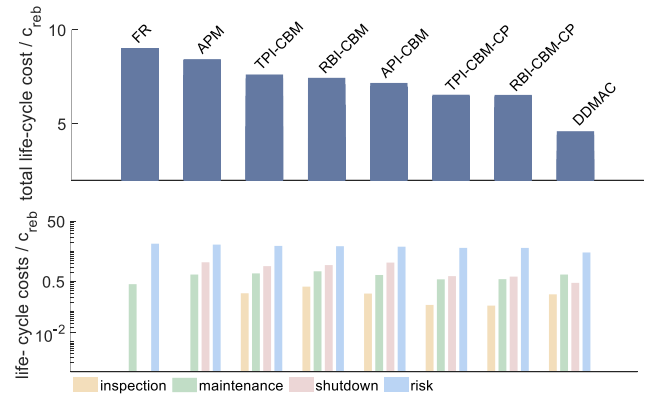

**Fig. 5**. Comparison of DDMAC life-cycle policies with different baseline policies. Total life-cycle cost and life-cycle costs due to inspection, maintenance, shutdown, and risk. The best optimized baseline is 42% worse than the DDMAC policy.

for component partial-repair and periodic age for component restoration/replacement.

- Age-Periodic Inspections and Condition-Based Maintenance (API-CBM) policy. Age-based inspections are taken for all components, based on each component's age. At inspection times, maintenance actions are taken based on the observed damage state of each component. Five variables are optimized; age interval for component inspection, and type of maintenance for each one of the 4 observed damage states.
- Time-Periodic Inspections and Condition-Based Maintenance (TPI-CBM) policy. Time-based inspections are taken for all components at fixed intervals of the planning horizon. At inspection times, maintenance actions are taken based on the observed damage state of each component. Five variables are optimized; time interval for block component inspection, and type of maintenance for each one of the 4 observed damage states.
- Risk-Based Inspections and Condition-Based Maintenance (RBI-CBM) policy. Inspections are taken for all components each time the system exceeds a predefined failure probability threshold. At inspection times, maintenance actions are taken based on the observed damage state of each component. Five variables are optimized; failure probability threshold, and type of maintenance for each one of the 4 observed damage states.

The last two baseline policies are also optimized with Component Prioritization (CP), which produces policies RBI-CBM-CP and TPI-CBM-CP. Components are prioritized based on their probability of failure. In this case, one extra decision variable regarding the number of components (1 to 10) to inspect and maintain is added. This policy adapts a heuristic presented in [10]. In all baselines, if a component fails, it is immediately replaced.

In Fig. 5, a comparison of the learned DDMAC policy with the various baselines is presented, for the unconstrained environment (total costs and decomposed costs in linear and log scales, respectively). The best optimal baseline is the policy combining risk-based inspections, condition-based maintenance and component prioritization. It can be observed that the life-cycle cost attained by the best baseline is about 42% worse than the DDMAC solution. The optimal age-periodic maintenance and fail-replacement policies, do not include the possibility of inspections and achieve the worst life-cycle costs. It is overall observed that baselines including inspections achieve consistently better results. Adding to this remark, it is interesting to note that the DDMAC policy spends more for inspections, i.e. performs a higher number of inspections, compared to the 3 best optimal baselines. As discussed, these inspections are in principle non-periodic and, as shown in Section 2.4, are driven by the innate notion of VoI in POMDPs. This allows the agents to make more informed decisions regarding proper maintenance actions that overall minimize the total cumulative costs of Eq. (19) more efficiently. Risk is significantly reduced with the DDMAC policy, as also indicated in Fig. 5, whereas scheduled system shutdown costs are more intelligently avoided compared to other baselines, due to the flexibility in intervention timings and action combinations.

### 4.4. Constrained system solutions

Constrained DDMAC results for life-cycle inspection costs, maintenance costs, shutdown costs and risk for different 5-year constraint levels are shown in Fig. 6 (all costs in log scale). As expected, higher budget limits result in lower total life-cycle costs. Budget limits higher than 25% of the system rebuild cost, $c_{reb}$, practically converge to the unconstrained solution. A noticeable feature of the learned near-optimal policies is that as budget becomes tighter, the agents tend to reduce their inspection expenses, to save resources in case of a need for major interventions (e.g. restoration/replacement actions). This means that they deliberately choose to forfeit better system information, in order to be more effective against disruption. It is characteristic that inspections are overall reduced in the budget cases below 15% $c_{reb}$, compared to the cases above that budget threshold, since the component replacement cost is 10% $c_{reb}$. That is, below 10% $c_{reb}$ budget constraints, restorations/replacements are infeasible. In Fig. 7, the respective results for risk constraints are shown (all costs in log scale). It can be observed that as the decision-making task becomes more risk averse the total life-cycle cost becomes higher, since more frequent inspection and maintenance actions need to be taken. Constrained solutions practically converge to the unconstrained one after the risk tolerance threshold of 2.75$c_{reb}$. It is interesting to note here that for lower risk constraints, i.e. for scenarios where the agents need to keep total risk lower over the operating life, although the maintenance cost increases, the inspection cost is not following the same trend, hence, the inspection per maintenance cost ratio of the optimal policy consistently decreases. This is attributed to the fact that more frequent maintenance is unavoidable in a case where risks have to be kept low, something that, by itself, leads on average, to longer sojourn in states of lower damage. As such, increased frequency of inspections, which would solely serve better state determination, is not favored by the agents, and thus life-cycle inspection costs do not present important changes for different risk-based constraints. Accordingly, due to the high demand for maintenance actions, scheduled shutdown costs also increase in low-risk cases.

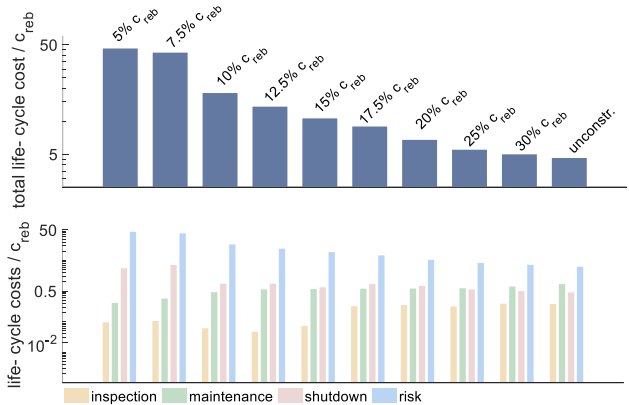

**Fig. 6**. Comparison of DDMAC life-cycle policies for different 5-year constraints from 5% $c_{reb}$ to infinity. Total life-cycle cost and life-cycle costs due to inspection, maintenance, shutdown, and risk.

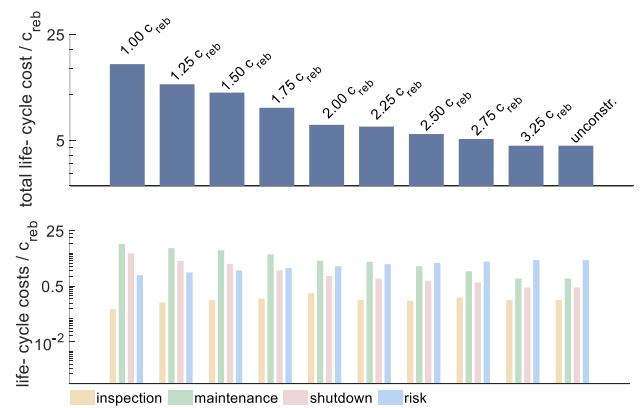

**Fig. 7**. Comparison of DDMAC life-cycle policies for different life-cycle risk constraints from 1 $c_{reb}$ to infinity. Total life-cycle cost and life-cycle costs due to inspection, maintenance, shutdown, and risk.

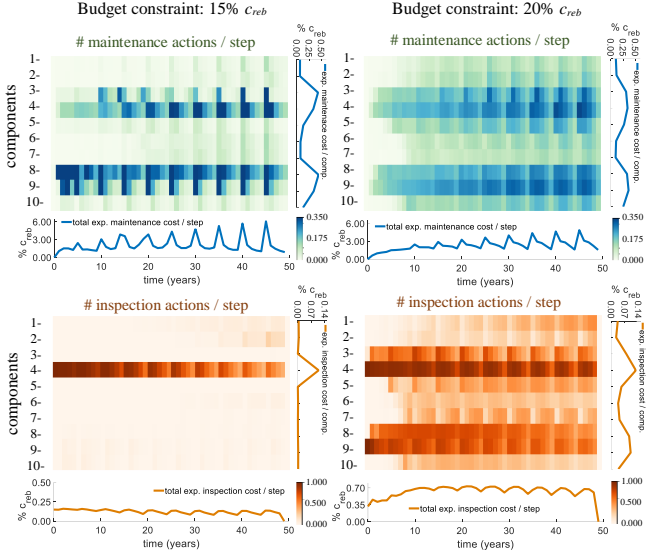

**Fig. 8**. Components maintenance and inspection frequency per step and respective mean costs for 5-year budget constraints of 15% and 20% $c_{reb}$.

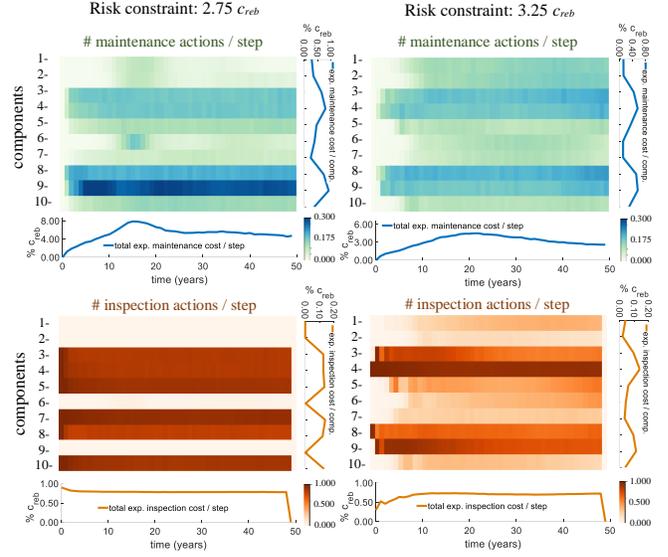

**Fig. 9**. Components maintenance and inspection frequency per step and respective mean costs for risk constraints of 2.75 and 3.25 $c_{reb}$.

In Fig. 8, action frequencies and respective cost metrics of inspection and maintenance are depicted for two budget constraints corresponding to a low and a high budget scenario, i.e. to 15% and 20% $c_{reb}$ 5-year budget constraints, respectively. Contour plots depict the frequency of maintenance and inspection actions per time unit. Adjacent graphs on the right show the mean step cost per component related to the respective action type, whereas the bottom graphs show the action cost per step, collectively for the all system components. The same features are depicted for risk constraints of 2.75 and 3.25 $c_{reb}$ in Fig. 9. Examining Figs. 6 and 8 together, we can observe that lowering the budget from 20% to 15% $c_{reb}$ has significant consequences for risk, which increases disproportionally with the achieved reduction in the expected total life-cycle maintenance cost. What changes significantly for maintenance cost, as shown in Fig. 8, is its distribution per time unit and component, rather than its total life-cycle value. This is indicative of the general observation that stricter budgets increase risk, without necessarily yielding clear economic budget-related benefits, if any, in the long run. Another interesting feature is that, in the presence of stricter budgets, the imbalance in the allocation of maintenance resources among components increases. Inspections and their respective expenditures are considerably restricted, as also mentioned previously. As also shown in Fig. 8, for the 15% $c_{reb}$ case, inspections are rather reserved mainly for component 4, as this is the most vulnerable component of path 6,7,4,5, which is the path securing system survival with the least number of components.

For the cases of risk-based constraints, examining Figs. 7 and 9 together, we can observe that relevant costs are distributed more evenly in time over the planning horizon. Over the system life-cycle, we observe that lowering the risk tolerance considerably encumbers maintenance costs per step and in total. Similarly to the budget-constrained cases, for the 2.75 $c_{reb}$ versus 3.25 $c_{reb}$ risk constraint case, inspections are prominently clustered to fewer components. Accordingly, it is observed that the agents reserve their inspection actions exclusively for components 3-5,7,8,10. This intrinsically prioritized selection of components to be frequently inspected allows the agents to track the state of at least half of the components from each link, and thereby to better synchronize maintenance actions in order to minimize system shutdowns and costs. It was observed that although mathematically feasible from an optimization perspective, policies below 2.0 $c_{reb}$ start becoming practically unrealistic due to the very frequent restorations/replacements that need to be taken in order for the risk constraints to be satisfied.

To look closer into how policies change for different constraints, 4 detailed policy realizations are shown in Figs. 10 and 11, for the constrained environments shown in Figs. 8 and 9, respectively. In Fig. 10(a), displaying the realization of component failure probabilities and respective inspection and maintenance actions, for two cases of 5-year budget constraints, it can be readily observed that, in the low budget scenario, available budgetary resources are primarily allotted to the maintenance needs of components 3,4,8, and 9. This is explained by the fact that these are Type III components, thus being described by the most aggressive deterioration. In this realization example, only component 4 is inspected, since, as also explained earlier, with a budget limit close to the component replacement cost, the agents choose to inspect more rarely in order to save resources if major interventions are needed. In the high budget scenario, inspections play a more prominent role, since the imposed

*C. P. Andriotis, K. G. Papakonstantinou / Deep reinforcement learning driven inspection and maintenance planning under incomplete information and constraints*

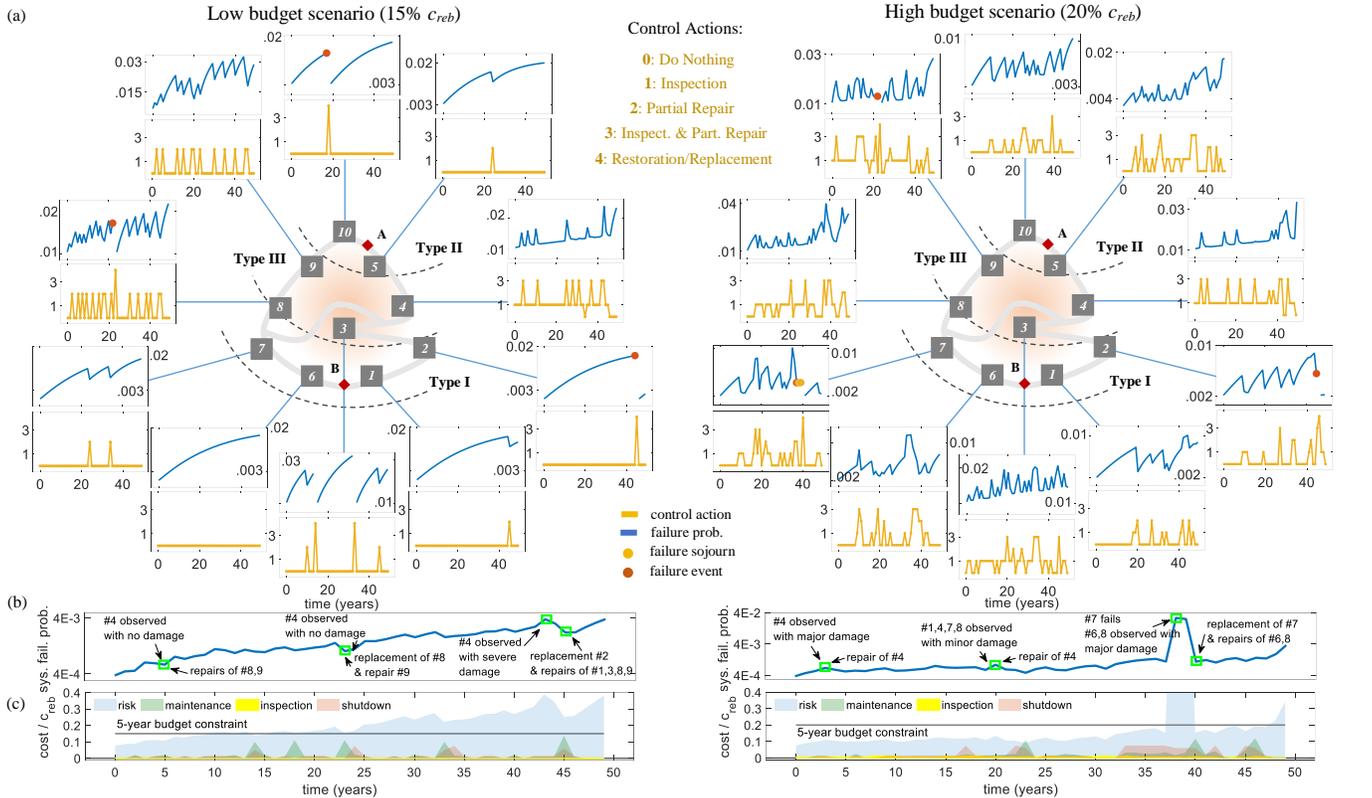

**Fig. 10**. A life-cycle realization of the DDMAC policy for 15% $c_{reb}$ and 20% $c_{reb}$ 5-year budget constraints. (a) Component failure probabilities and actions; (b) System failure with selected interventions; (c) Costs of inspection and maintenance actions, scheduled shutdowns, and risks.

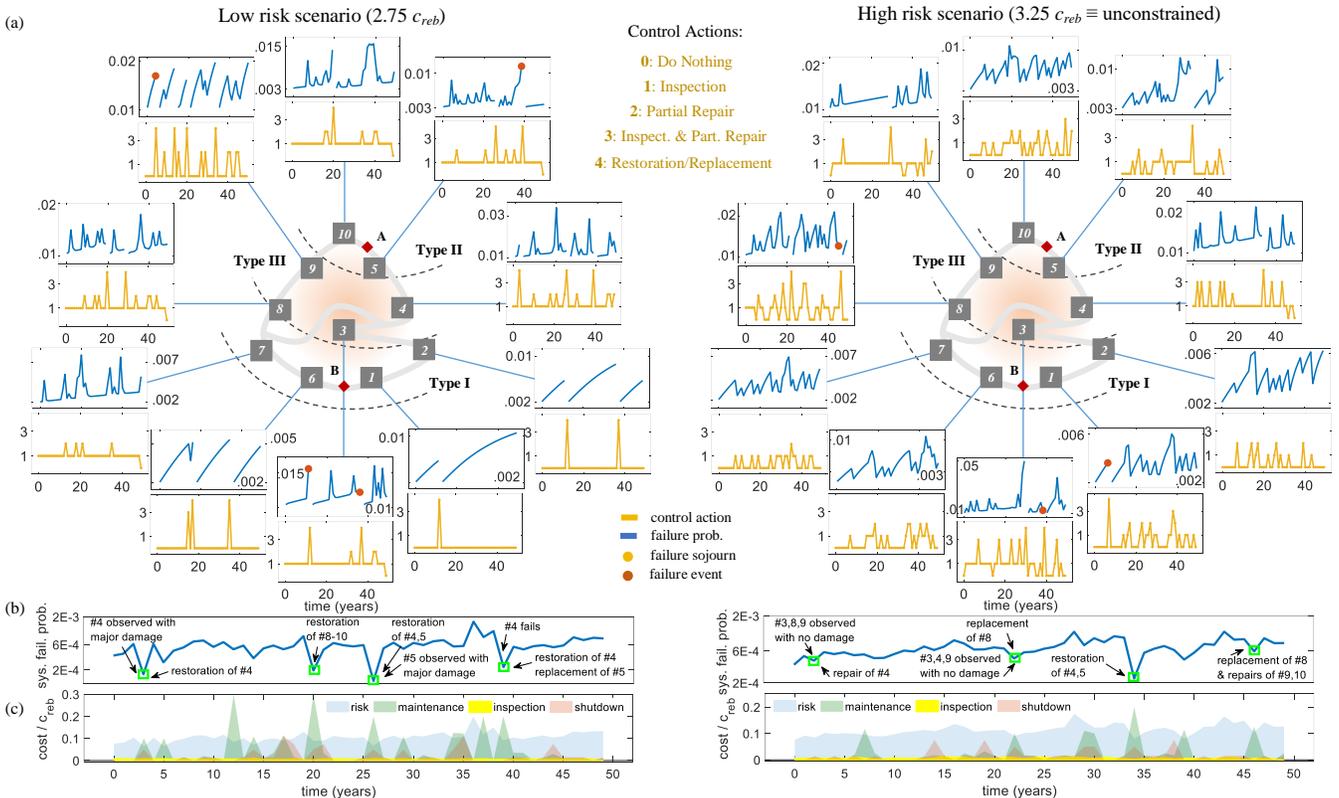

**Fig. 11**. A life-cycle realization of the learned DDMAC policy for 2.75 $c_{reb}$ and 3.25 $c_{reb}$ life-cycle risk constraints. (a) Component failure probabilities and actions; (b) System failure probability with selected interventions; (c) Costs of inspection and maintenance actions, scheduled shutdowns, and risks.

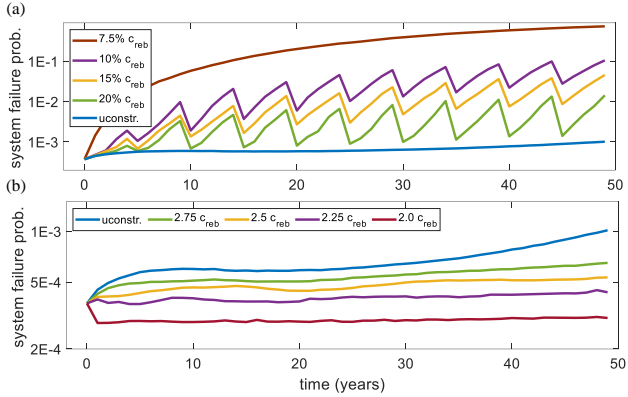

**Fig. 12**. System mean failure probabilities based on DDMAC life-cycle policies for different (a) 5-year budget constraints and (b) life-cycle risk constraints.

budget restrictions have become looser, and the agents have the budgetary capacity to afford expenditure for acquiring information. Although Type III components continue to receive the majority of maintenance actions, intervention resources are now allotted more frequently to all components. Some of the most prominent intervention effects changing significantly the overall system failure probability, are indicated in Fig. 10(b). The various costs are also tracked in Fig. 10(c). For the 20% $c_{reb}$ case, a notable feature can be observed for components 6 and 7, controlling the operability of the third link. Component 7 fails at $t=38$ and available resources do not allow for immediate replacement, which is postponed to $t=40$, when the next budget cycle begins. In the meanwhile, the agent of component 6 takes advantage of the link shutdown and applies repeated opportunistic partial repairs which do not yield additional shutdown costs. Overall, it can be interestingly observed in Figs. 8 and 10, that the agents, despite their decentralized policies, form and increase collaboration as the budget becomes lower, directing their focus to components that are more vulnerable to deterioration, or more strategically placed in terms of system connectivity.

Similar features can be seen for the low- and high-risk constraints cases of Fig. 11. In the 3.25 $c_{reb}$ case, effectively coinciding with the unconstrained policy, a complex and diverse policy is overall illustrated. It is worth noting that, in the absence of any budget constraints, inspections are now taken frequently for all components, whereas restoration/replacement actions start to also have more prominent preventive characteristics, i.e. they are not only reserved for failure events. This is even more apparent in the low-risk scenario, in which case, restorations need to be performed in a more recurrent fashion to ensure low probability of failure. In turn, this also causes more system closures and, thus, increases shutdown costs. To balance this side effect of frequent restorative actions, the agents are interestingly shown to deploy a block-restoration/replacement logic in their policies. That is, as shown in the 2.75 $c_{reb}$ scenario of Fig. 11(a), component agents of the same links synchronize their restoration actions (e.g. components 2,3 at $t=37$; components 8-10 at $t=20$; components 4,5 at $t=26$), whereas they also start to extensively leverage opportunistic interventions in links where failure events occur (e.g. components 1-3 at $t=12$; components 4,5 at $t=39$). The system failure probability and the various costs along with various actions that affect them are shown in Figs. 11(b),(c), respectively.

The mean failure interval probability of the system over time is shown in Fig. 12, for various 5-year budget and various life-cycle risk constraints. It is observed that, on average, system failure probability reaches its peak before the onset of new budget cycles. For the unconstrained case, mean failure probability is allowed to increase over time, without abrupt escalations, since no budget limitation is imposed. The 7.5% $c_{reb}$ constrained case reflects an extreme life-cycle optimization setting where no replacement actions are feasible. Thus, in this case no major corrective steps are detected in the evolution of the mean failure probability. In the case of risk constraints, the more stringent the risk constraint is, the higher is the reliability of the system at each time step, as anticipated.

## 5. Conclusions

In this work a stochastic optimal control framework for inspection and maintenance planning of deteriorating systems operating under incomplete information and constraints is developed. Decision-making is cast in a multi-agent decentralized framework of DRL and POMDPs, where each system component, or control unit consisting of multiple components, acts as an independent agent given the dynamically updated global system state probabilistic information. While satisfying a shared overarching objective, each agent can make its own inspections and maintenance choices. Operational resource-based restrictions and policy risk considerations are taken into account by means of relevant stochastic soft and/or hard constraints. The latter are incorporated in the solution scheme through state augmentation, thus being rendered as environment properties, whereas the former are appended in the life-cycle objective function as dual variables, to form the Lagrangian function to be optimized. Modeling of various constraint choices is discussed, whereas a thorough numerical investigation is provided for budget and risk constraints, which are of particular significance in infrastructure management applications. Along these lines, a broad risk definition is also presented and utilized in the constrained optimization procedure, accommodating both the instantaneous and perpetual nature of damage-related losses. This risk definition is further shown to be reducible to classic reliability-based definitions. Solutions to the optimization problem are driven by the introduced DDMAC algorithm. DDMAC uses both policy and value function parametrizations, experience replay, off-policy network parameter updating, and operates on the belief space of the underlying POMDP.

Operation under constraints is shown to considerably affect how the agents adapt their policies. The conducted parametric analysis shows that:

- The need for inspections and, therefore, the value of information, fades in low-budget environments, where the agents tend to diminish expenses otherwise allotted to system updating needs, in order to secure advanced intervention capabilities through availability of maintenance resources.
- Stricter budget constraints reduce inspection and maintenance costs for the respective budget cycle, however, without comparably reducing these costs in the long run, i.e., cumulatively, over the system life-cycle.
- In risk-averse environments, inspection costs do not follow the notable increase in maintenance costs, which are necessary in order to maintain low-risk levels over the system operating life.

- In such cases, agents are shown to increasingly leverage the structural properties of the system or incidental sub-system failure configurations, to develop opportunistic repair strategies, so that system operability is minimally disrupted.
- Budget limitations and risk intolerance, disproportionally increase the risk and maintenance costs, respectively, compared to the reductions they achieve in the constrained quantities.
- For both types of constraints, multi-agent cooperation emerges more prevalent as restrictions become stricter, since resource scarcity and risk intolerance force the agents to more carefully reallocate resources and redefine management priorities, based on the specific deterioration dynamics and structural importance of different system parts. This rescheduling arises naturally and intrinsically through the training process, without any explicit user-based enforcement or penalty-driven motivation.

Overall, the derived DRL policies showcase remarkable flexibility and multi-agent cooperation in various constrained and unconstrained environments, whereas the obtained decentralized solutions are found to significantly outperform conventional and state-of-the-art inspection and maintenance planning formulations.

## Acknowledgements

This material is based upon work supported by the U.S. National Science Foundation under CAREER Grant No. 1751941, and the Center for Integrated Asset Management for Multimodal Transportation Infrastructure Systems (CIAMTIS), 2018 U.S. DOT Region 3 University Center.

## Appendix A: On the definition of risk

**Proposition A.1**. If only failure incurs damage cost, this cost is $c_F$, and it is instantaneous, then:

$$\mathfrak{R}^\pi = \mathbb{E}_{o_{0:T}}\left[\sum_{t=0}^{T}\gamma^t \mathbb{E}_{s_t^a, s_{t+1}}\left[c_D^{per}(s_{t+1}) + d_{s_t^a s_{t+1}} c_D^{inst}(s_{t+1})\right]\right]$$

$$= c_F \mathbb{E}_{o_{0:T}}\left[\sum_{t=0}^{T}\gamma^t \left(P_{F_{t+1}|a_{0:t},o_{0:t}} - P_{F_t|a_{0:t},o_{0:t}}\right)\right] = \mathfrak{R}_F^\pi$$

where $P_{F_{t+1}|a_{0:t},o_{0:t}}$ and $P_{F_t|a_{0:t},o_{0:t}}$ are the probabilities of failure up to time $t+1$ and $t$, respectively, given a history of actions and observations $a_{0:t}$, $o_{0:t}$.

*Proof.* By definition, the transition probability from $s^a$ to $s'$ can be written as:

$$\Pr(s'|s^a,a) = (d_{s^a s'} + \delta_{s^a s'})\Pr(s'|s^a,a) \quad (A.1)$$

where $d_{ij}$ and $\delta_{ij}$ are the adjacency and Kronecker indicators, as defined in Eqs. (14) and (7), respectively, for all $i,j$ belonging to $S$. Thus, using Eq. (A.1), the expected cost of the instantaneous costs of Eq. (14) reads:

$$c_{b,D}^{inst} = \mathbb{E}_{s^a,s'}\left[d_{s^a s'} c_D^{inst}(s')\right]$$
$$= \mathbb{E}_{s^a,s'}\left[(d_{s^a s'} + \delta_{s^a s'})(1 - 1 + d_{s^a s'})c_D^{inst}(s')\right]$$

$$= \mathbb{E}_{s^a,s'}\left[c_D^{inst}(s') - (1 - d_{s^a s'})(d_{s^a s'} + \delta_{s^a s'})c_D^{inst}(s')\right]$$
$$= \mathbb{E}_{s^a,s'}\left[c_D^{inst}(s') - \delta_{s^a s'} c_D^{inst}(s')\right]$$
$$= \mathbb{E}_{s^a,s'}\left[c_D^{inst}(s')\right] - \mathbb{E}_{s^a,s'}\left[\delta_{s^a s'} c_D^{inst}(s')\right]$$
$$= \sum_{s^a \in S} b^a(s^a) \sum_{s' \in S} \Pr(s'|s^a,a) c_D^{inst}(s')$$
$$\quad - \sum_{s^a \in S} b^a(s^a) \Pr(s'=s^a|s^a,a) c_D^{inst}(s^a) \quad (A.2)$$

Combining both instantaneous and perpetual damage costs, and elaborating further on Eq. (15) we finally obtain:

$$c_{b,D} = \mathbb{E}_{s^a,s'}\left[c_D^{per}(s') + c_D^{inst}(s')\right] - \mathbb{E}_{s^a,s'}\left[\delta_{s^a s'} c_D^{inst}(s')\right]$$
$$= \sum_{s^a \in S} b^a(s^a) \sum_{s' \in S} \Pr(s'|s^a,a)\left(c_D^{per}(s') + c_D^{inst}(s')\right)$$
$$\quad - \sum_{s^a \in S} b^a(s^a) \Pr(s'=s^a|s^a,a) c_D^{inst}(s^a) \quad (A.3)$$

Using Eq. (A.3), risk is defined as the cumulative damage state cost over the life-cycle in expectation:

$$\mathfrak{R}^\pi = \mathbb{E}_{o_{0:T}}\left[\sum_{t=0}^{T}\gamma^t \left(\mathbb{E}_{s_t^a, s_{t+1}}\left[c_D^{per}(s_{t+1}) + c_D^{inst}(s_{t+1})\right]\right.\right.$$
$$\left.\left. - \mathbb{E}_{s_t^a, s_{t+1}}\left[\delta_{s_t^a s_{t+1}} c_D^{inst}(s_{t+1})\right]\right)\right] \quad (A.4)$$

Eq. (A.4) is equivalent to Eq. (15). We now consider that only failure incurs damage related cost, this cost is $c_F$, and it is instantaneous. We denote failure state(s) as $s_F$. The respective cost is written as $c_D^{inst} = \delta_{s_{t+1} s_F} c_F$. In this case, Eq. (A.4) reduces to:

$$\mathfrak{R}^\pi = \mathbb{E}_{o_{0:T}}\left[\sum_{t=0}^{T}\gamma^t \left(\mathbb{E}_{s_t^a, s_{t+1}}\left[\delta_{s_{t+1} s_F} c_F\right] - \mathbb{E}_{s_t^a, s_{t+1}}\left[\delta_{s_t^a s_{t+1}} \delta_{s_{t+1} s_F} c_F\right]\right)\right]$$
$$= c_F \mathbb{E}_{o_{0:T}}\left[\sum_{t=0}^{T}\gamma^t \left(\mathbb{E}_{s_t^a, s_{t+1}}\left[\delta_{s_{t+1} s_F}\right] - \mathbb{E}_{s_t^a}\left[\delta_{s_t^a s_F}\right]\right)\right]$$
$$= c_F \mathbb{E}_{o_{0:T}}\left[\sum_{t=0}^{T}\gamma^t \left(\sum_{s_t^a \in S} b^a(s_t^a) \sum_{s_{t+1} \in S} \Pr(s_F|s_t^a, a_t) - b^a(s_F)\right)\right] \quad (A.5)$$

Probability $b^a(\cdot)$, is the updated probability, as defined by Eqs. (4) and (6), hence, $b^a(\cdot) = \Pr(\cdot|a_{0:t}, o_{0:t})$. As such, Eq. (A.5) is equivalently written as:

$$\mathfrak{R}^\pi = c_F \mathbb{E}_{o_{0:T}}\left[\sum_{t=0}^{T}\gamma^t \left(P_{F_{t+1}|a_{0:t},o_{0:t}} - P_{F_t|a_{0:t},o_{0:t}}\right)\right] \quad (A.6)$$

From (A.6) follows immediately that $\mathfrak{R}^\pi = \mathfrak{R}_F^\pi$. □

Under the assumptions of the above proposition, one can model the POMDP problem with damage cost:

$$c_D(s,a,s^a,s') = d_{s^a s'} \delta_{s' s_F} c_F = \delta_{s' s_F} c_F - \delta_{s^a s_F} c_F \quad (A.7)$$

Marginalizing with respect to $s'$ and assuming that pair $(s,a)$ leads deterministically to $s^a$, a simpler expression can also be used for the cost function:

$$c_D(s,a) = \Pr(s_F \mid s^a, a)c_F - \delta_{s^a s_F} c_F \tag{A.8}$$

Note that Eq. (A.8) is a closed form expression, if transitions to failure state from all other states, $\Pr(s_F \mid s^a,a)$, are known, i.e. according to standard offline Markov decision processes semantics.